# Updating Industrial Robots for Emerging Technologies

**David Puljiz[1], Björn Hein[2],**
[1] **Karlsruhe Institute of Technology,**
david.puljiz@kit.edu
[2] **Karlsruhe University of Applied Sciences,**
bjoern.hein@kit.edu

Industrial arms need to evolve beyond their standard shape to embrace new and emerging technologies. In this paper, we shall first perform an analysis of four popular but different modern industrial robot arms. By seeing the common trends we'll try to extrapolate and expand these trends for the future. Here, particular focus will be on interaction based on augmented reality (AR) through head-mounted displays (HMD), but also through smartphones. Long-term human-robot interaction and personalization of said interaction will also be considered. The use of AR in human-robot interaction has proven to enhance communication and information exchange. A basic addition to industrial arm design would be the integration of QR markers on the robot, both for accessing information and adding tracking capabilities to more easily display AR overlays. In a recent example of information access, Mercedes Benz added QR markers on their cars to help rescue workers estimate the best places to cut and evacuate people after car crashes. One has also to deal with safety in an environment that will be more and more about collaboration. The QR markers can therefore be combined with RF-based ranging modules, developed in the EU-project SafeLog, that can be used both for safety as well as for tracking of human positions while in close proximity interactions with the industrial arms. The industrial arms of the future should also be intuitive to program and interact with. This would be achieved through AR and head mounted displays as well as the already mentioned RF-based person tracking. Finally, a more personalized interaction between the robots and humans can be achieved through life-long learning AI and disembodied, personalized agents. We propose a design that not only exists in the physical world, but also partly in the digital world of mixed reality.

## Introduction

Industrial robot design needs to evolve with the changing technological and social landscape. In particular, the design of industrial arms has barely changed since their first introduction more than 65 years ago. The relatively recent development of collaborative robots being the only exception to this rule. The current drive is for more collaborative robots integrating new safety technology and design rules into industrial arms. In particular the introduction of joint-torque sensors for safety but also for enabling more intuitive programming through hand guidance. Other technologies such as touch sensitive and soft skins have also been introduced. Finally, due to the ever-increasing proximity of the interaction between humans and robots, industrial arms are required to appear friendlier and safer. However, as trends for better safety sensors, more intuitive programming, human robot collaboration and indeed miniaturization continue, there is no drive to go beyond the well-established designs and technologies. For the authors, perhaps the most exciting and important emerging technology is the Head-Mounted Display

(HMD) and in parallel to it Augmented Reality (AR). Both of these can be used to greatly improve interaction, programming and safety in human-robot interactions (HRI), due to more intuitive information display, localization and mapping capabilities, head tracking, and other useful sensors. The use of AR also allows us to put part of the design into the virtual world, visible through the use of HMDs but also through smartphones.

## Analysis of Current Desings

We shall compare the design elements of four different industrial arms. One of which (Kuka KR50) is a standard industrial robot arm not inherently designed for collaboration. The other three are collaborative robots (Baxter, FANUC CR-35iA and the UR-16e). The design features that were evaluated were taken from the technical specifications and online material. In Fig. 1 we have detailed the common design elements as well as the peculiarities of each robot. In Table 1 we list the design features of each model alongside the desired effects. One can immediately see that the Kuka KR50 arm was built with efficiency in mind and not intended for proximal HRI, at least not without heavy modification. There are no features to ease programming and no in-built safety features to protect a human coworker. On the other hand, such arms are efficient and have higher payloads than comparable collaborative robots. The collaborative robots all have in-built joint-torque sensors. These sensors can be used to detect collisions and limit the force of impact, ensuring the safety of the human coworker. Additionally, they can be used to program the arm through hand guidance, also called lead-through programming. The robot arm can be programmed by simply grabbing the robot arm and showing it the trajectory to perform. Hand guidance can also be used to intervene mid-execution to adapt on the fly. Collaborative robots are designed to be less intimidating and more "friendly". For example the

| Robot Model | Design Features | Effects |
|---|---|---|
| KR50 | 50 kg high dynamic carry weight | Mechanically efficient, large carry weight |
| Baxter | 2.2 kg carry weight, joint-torque sensors for safety in HRI, screen with default friendly face | Safe for HRI, positive psychological effect, better force control, easier programming through lead-through method |
| CR-35iA | 35 kg carry weight, Joint-torque sensor in base and rubber skin for safety in HRI | Safe for HRI, positive psychological effect |
| UR-16e | 16 kg carry weight, Joint-torque sensors in every joint for safety. Sensors also enable better force control | Safe for HRI, better force control, easier programming through lead-through method |

**Table 1. Design features of different industrial arm models and relevant effects**

Baxter robot incorporates a screen with a friendly digital face, while the CR-35iA, although quite imposing, has a rubber exterior as an additional layer of protection. The CR-35iA only possesses a force sensor in the base, meaning that an additional module is needed for hand guidance. It can also lift up to 35kg, making it comparable to standard industrial arms. As we envision human-robot collaboration to be paramount for future industrial manipulators, the design elements of the collaborative robots must be adopted. Many of the more complex elements of such designs, such as the rubber skin, joint-torque sensors, the "friendliness" and programming through hand guidance may be replaced through the use of HMDs, AR and RF safety sensors.

## Proposed Future Design

The proposed new design combines all of the elements discussed in the previous section into a single streamlined robot. The design can be seen in Fig. 2. The basis of the industrial arm would be the standard, efficient industrial robot design similar to the Kuka KR50. To achieve functional safety, RF modules would be installed on the surface of the robot, at least two for each link, as well as carried by the human, as watches on the wrist and in the form of a vest for the body. The sensors were developed for functional safety in mind for the project SafeLog by the partners at Končar and UNIZG-FER. The working principle of these sensors is outlined in [1]. In this scenario they would stop the robot arm if any part gets too close to the human as well as provide distances between the human-worn sensors and the sensors mounted on the robot arm. This would cover all of the safety features of the current designs while also allowing the robot to know exactly where the human is.

Adding QR markers on the RF nodes may be used to gain information about the robot system and troubleshoot problems on the fly. The QR markers can also be used for tracking and displaying AR information through the use of HMDs. To note is that any QR marker on a fast-moving link of the manipulator will suffer from motion blur when captured through the HMD's camera, rendering it unusable for tracking. This is mitigated by the redundancy of the markers and the markers on the robot base itself that always provide a baseline.

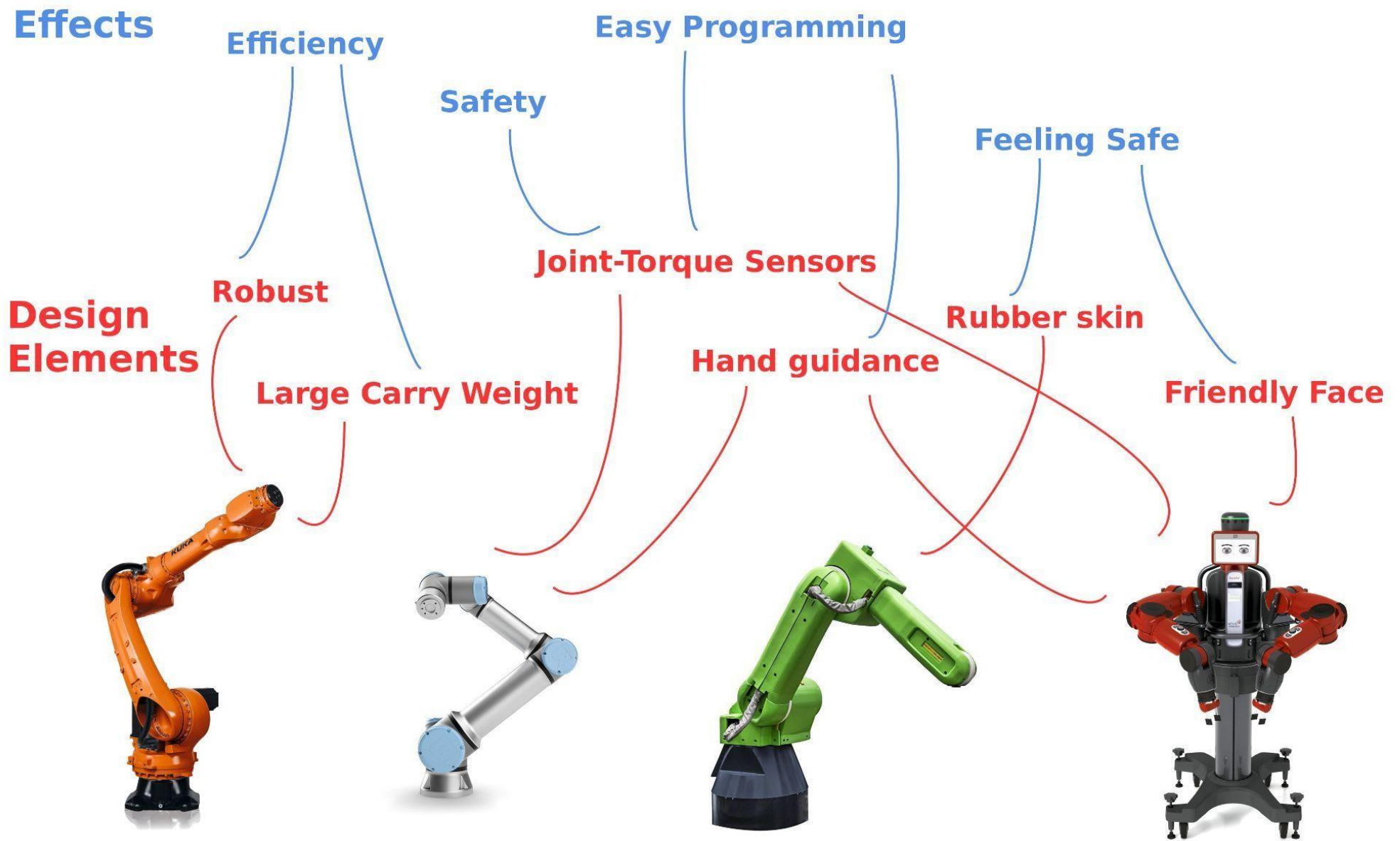

**Figure 1. Design analysis of common industrial robot arms - (from left to right) Kuka KR-50, Universal Robotics' UR-16e, FANUC CR-35iA, Rethink Robotics' Baxter**

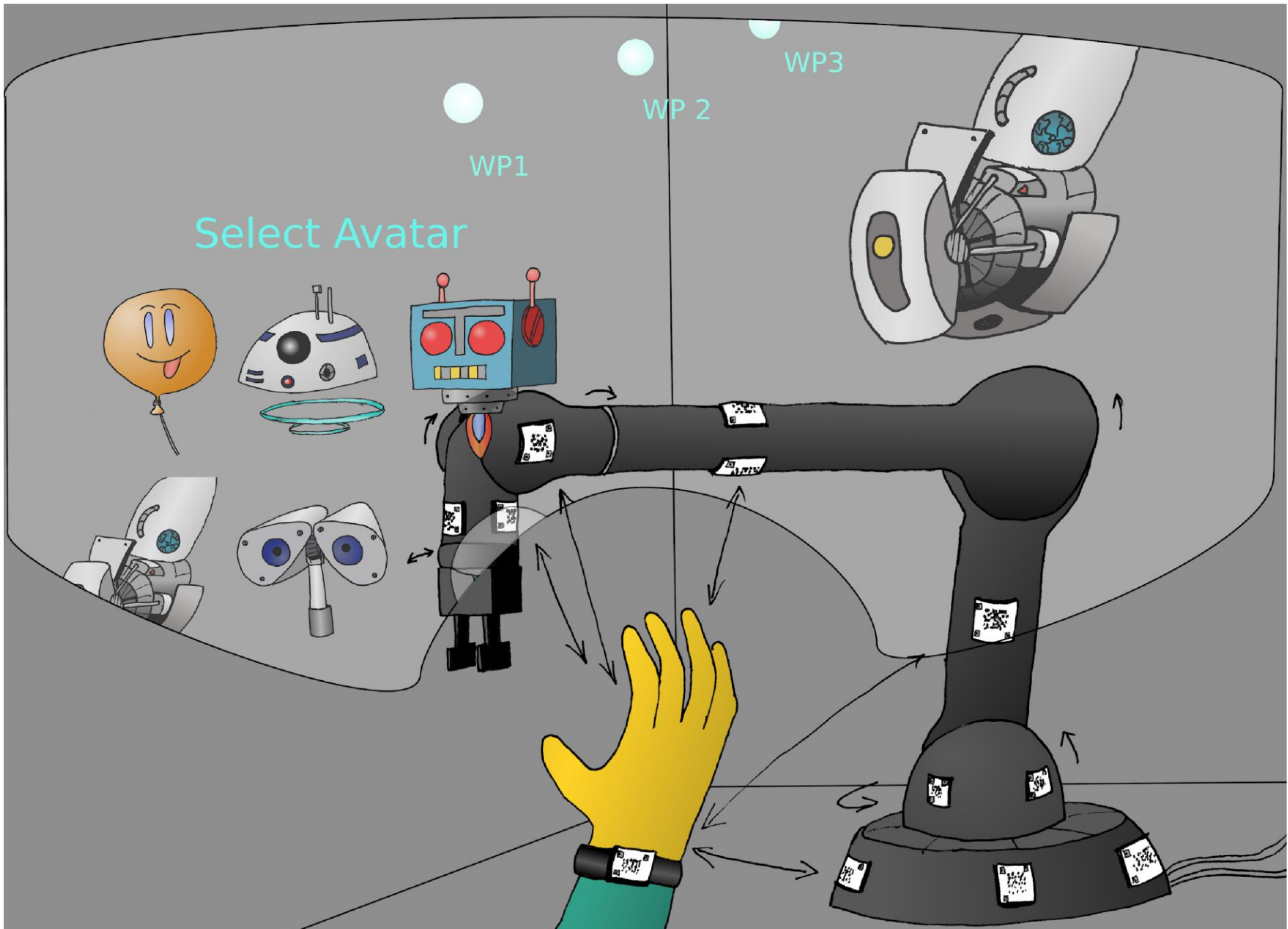

**Figure 2. Proposed design of future industrial robot arms including RF distance modules with QR markers, AR interaction and virtual avatars. Note the darker colors**

Through the use of the RF modules, the localization and hand tracking capabilities of the HMD, one can recreate the hand-guidance programming of collaborative robots quite effectively, as was shown in [2]. In addition, novel methods of programming, such as adding holographic waypoints, or being able to track programming tools through the HMD open up a vast array of intuitive programming methods not available to current systems. The HMD can also be used to map the environment of the industrial arm, making set-up fast and easy, as illustrated by our work [3].

The arm, unlike most current designs, would be black. The color black in particular is quite appropriate for HMDS, as they add light but cannot subtract it and, therefore, have problems with overlaying AR information on top of bright surfaces. On the other hand, if one considers the use of infrared depth sensors for spatial mapping, the black color proves to be a hindrance as the depth sensors prefer reflective surfaces. A surprising benefit of the manipulator being "invisible" to depth sensors is when it comes to workspace mapping. To get a usable map however, the robot itself needs to be removed from the map which is a relatively costly process. An invisible robot would remove the need for this computationally expensive step.

Finally, the personalization and friendliness aspect would be handled by customizable AR avatars. Each person can select their own avatar. This avatar would not only be visual, but would also have intelligence, similar to Alexa or Google Assistant. The virtual avatars would learn preferences and personalities of each human coworker through the novel methodologies of lifelong learning and long-term HRI concepts. They could also be endowed with their own personalities (sarcastic, happy, shy etc.). We believe that the inclusion of such avatars would enhance the trust in human-robot teams and the safety the human coworker feels while working with the robot companion. Furthermore, as the avatars are immaterial, as they are based on AR, they can be carried along and projected to any robot the human might be working with. The human may also choose a distinct avatar and "personality" for each robot coworker.

## Conclusion

We strongly believe that emerging technologies will have a strong influence on how we perceive and interact with robots. In this paper we applied some emerging technologies on the relatively outdated area of industrial robot arms. We have shown how all of the current trends can be kept and also expanded through the use of RF-based safety ranging, HMDs, AR interaction and smart, customizable avatars. We presented a design that not only exists in hardware and the physical world, but also to a great extent in the digital and mixed reality.

## Acknowledgement

This work was supported by the Federal Ministry of Education and Research of Germany within the Program Research for Civil Security Project ROBDEKON (Funding No. 13N14678)